\documentclass{article}
\usepackage{latexsym,amssymb,amsmath}
\usepackage{hyperref}
\usepackage{times}
\usepackage[accepted]{icml2016} 
\usepackage{path}
\usepackage{url}
\usepackage{amsmath}
\usepackage{subcaption}
\usepackage{tikz}
\usetikzlibrary{arrows}
\usepackage{verbatim}
\usepackage{circuitikz}
\usetikzlibrary{positioning}
\usepackage{tkz-graph}
\usepackage{fancyhdr}
\usepackage{booktabs}
\usepackage{courier}
\usetikzlibrary{decorations.markings}
\usetikzlibrary{shapes,snakes}
\usepackage[toc,page]{appendix}
\usepackage{verbatim}
\usepackage{alltt}
\usepackage{sverb}
\usepackage{listings}
\usepackage{color}
\usepackage{lipsum}
\usepackage{array}
\usepackage{mathtools}
\usepackage{subcaption}

\numberwithin{equation}{section}
\numberwithin{table}{section}
\numberwithin{figure}{section}

\definecolor{codegreen}{rgb}{0,0.6,0}
\definecolor{codegray}{rgb}{0.5,0.5,0.5}
\definecolor{codepurple}{rgb}{0.58,0,0.82}
\definecolor{backcolour}{rgb}{0.95,0.95,0.95}
\tikzstyle{block} = [draw,minimum size=2.5em, outer sep=2]

\begin{document}

\tikzset{->-/.style={decoration={
  markings,
  mark=at position #1 with {\arrow{>}}},postaction={decorate}}}


\twocolumn[
\icmltitle{Suprisal-Driven Zoneout}

\icmlauthor{Kamil Rocki}{kmrocki@us.ibm.com}
\icmlauthor{Tomasz Kornuta}{tkornut@us.ibm.com}
\icmladdress{IBM Research, San Jose, CA 95120, USA}
\icmlauthor{Tegan Maharaj}{tegan.maharaj@polymtl.ca}
\icmladdress{Ecole Polytechnique de Montreal}

\icmlkeywords{LSTM, recurrent, surprisal, wikipedia}

\vskip 0.3in
]


\begin{abstract}


We propose a novel method of regularization for recurrent neural networks called suprisal-driven zoneout. In this method, states \textit{zoneout} (maintain their previous value rather than updating), when the \textit{suprisal} (discrepancy between the last state's prediction and target) is small. Thus regularization is adaptive and input-driven on a per-neuron basis. We demonstrate the effectiveness of this idea by achieving state-of-the-art bits per character of 1.31 on the Hutter Prize Wikipedia dataset, significantly reducing the gap to the best known highly-engineered compression methods. 



\end{abstract}





\section{Introduction}





An important part of learning is to go beyond simple memorization, to find as general dependencies in the data as possible. For sequences of information, this means looking for a concise representation of how things change over time. One common way of modeling this is with recurrent neural networks (RNNs), whose parameters can be thought of as the transition operator of a Markov chain. Training an RNN is the process of learning this transition operator. Generally speaking, temporal dynamics can have very different timescales, and intuitively it is a challenge to keep track of long-term dependencies, while accurately modeling more short-term processes as well. 

The Long-Short Term Memory (LSTM) \citep{Hochreiter:1997:LSM:1246443.1246450} architecture, a type of RNN, has proven to be exceptionally well suited for learning long-term dependencies, and is very widely used to model sequence data. Learned, parameterized gating mechanisms control what is retrieved and what is stored by the LSTM's state at each timestep via multiplicative interactions with LSTM's state.
There have been many approaches to capturing temporal dynamics at different timescales, e.g. neural networks with kalman filters, clockwork RNNs, narx networks, and recently hierarchical multiscale neural networks. 







It has been proven \citep{solomonoff1964formal} that the most general solution to a problem is the one with the lowest Kolmogorov complexity, that is its code is as short as possible. In terms of neural networks one could measure the complexity of a solution by counting the number of active neurons. According to \emph{Redundancy-Reducing Hypothesis}~\cite{barlow1961possible} neurons within the brain can code messages using different number of impulses. This indicates that the most probable events should be assigned codes with fewer impulses in order to minimize energy expenditure, or, in other words, that the more frequently occuring patterns in lower level neurons should trigger sparser activations in higher level ones. Keeping that in mind, we have focused on the problem of adaptive regularization, i.e. minimization of a number of neurons being activated depending on the novelty of the current input. 

\begin{figure}[t!]
\centering 
\includegraphics[width=0.9\columnwidth]{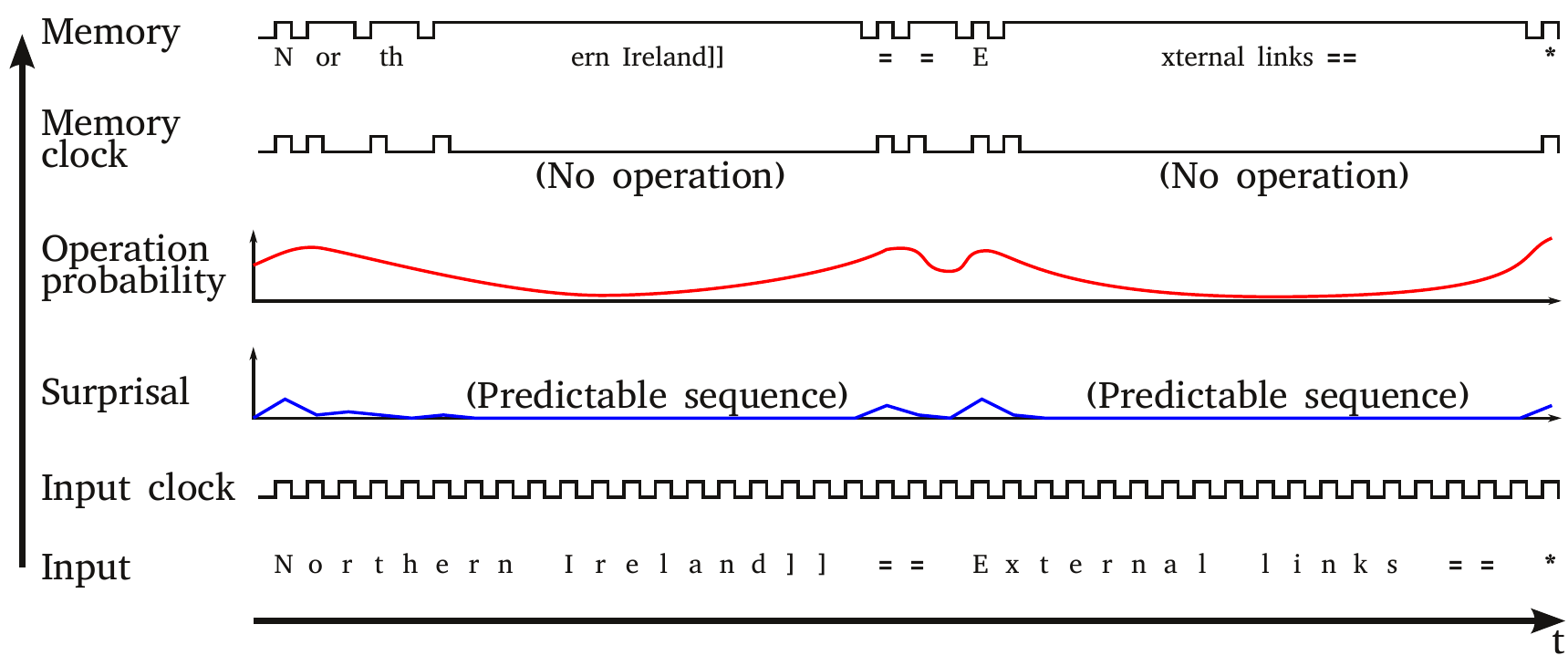}
\caption{Illustration of the adaptive zoneout idea}
\label{fig:idea}
\end{figure}


Zoneout is a recently proposed regularizer for recurrent neural networks which has shown success on a variety of benchmark datasets ~\citep{DBLP:journals/corr/KruegerMKPBKGBL16, rocki2016recurrent}.
Zoneout regularizes by \emph{zoning} out activations, that is: \emph{freezing} the state for a time step with some fixed probability. This mitigates the unstable behavior of a standard dropout applied to recurrent connections.
However, since the zoneout rate is fixed beforehand, one has decide a priori to prefer faster convergence or higher stochasticity, whereas we  would like to be able to set this per memory cell according to learning phase, i.e. lower initially and higher later to prevent memorization/unnecessary activation. This is why we have decided to add surpisal-driven feedback~\citep{rocki2016surprisal}, since it gives a measurement of current progress in learning.
The provided (negative) feedback loop enables to change the zoneout rate online within the scope of a given cell, allowing the zoneout rate to adapt to current information.
As learning progresses, the activations of that cell become less frequent in time and more iterations will just skip memorization, thus the proposed mechanism in fact enables different memory cells to operate on different time scales. 
The idea is illustrated in Fig. \ref{fig:idea}. 

The main contribution of this paper is the introduction of Surprisal-Driven Adaptive Zoneout, where each neuron is encouraged to be active as rarely as possible with the most preferred state being \textit{no operation}. The motivation behind this idea is that low complexity codes will provide better generalization.



\section{The model}

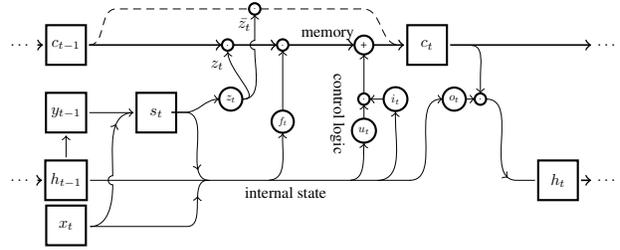
\begin{figure}[h]
\centering
\centering
\begin{tikzpicture}[thick,scale=0.6, every node/.style={scale=0.6}]
   \foreach \y [evaluate = \y as \expr using (\y*0.4-1.2), evaluate = \y as \exprr using (\y-3.5), evaluate = \y as \exprrr using  int(3-\y)]  in {0} {
   	   \node[] at (0,-\y) (input_prev\y) {$\dots$};
       \node[block] at (1,-\y) (input\y) {$c_{t-1}$};
       \node[draw,circle, xscale=0.6, yscale=0.6]at (5.8,-\y) (f_\y) {.};
        \node[draw,circle, xscale=0.7, yscale=0.7]at (7.6,-\y) (f2_\y) {+};
	 \node[draw,circle, xscale=0.65, yscale=0.65] at (-\expr+4.6,-1.7) (f_block\y) {$f_{t}$};
  \node[draw,circle, xscale=0.75, yscale=0.75] at (-\expr+6.4,-1.9) (i_block\y) {${u}_{t}$};
  \node[draw,circle, xscale=0.75, yscale=0.75] at (8.3,\expr) (g_block\y) {$i_{t}$};
  \node[draw,circle, xscale=0.75, yscale=0.75] at (9.6,\expr) (o_block\y) { $o_{t}$};
  \node[draw,circle, xscale=0.6, yscale=0.6] at (-\expr+6.4,\expr) (ig_block\y) {.};
  \node[draw,circle, xscale=0.6, yscale=0.6] at (\expr+11.38,\expr) (oc_block\y) {.};
  \node[draw,circle, xscale=0.6, yscale=0.6] at (\expr+6.38,\expr+2) (s0_\y) {.}; 
  \node[draw,circle, xscale=0.6, yscale=0.6] at (\expr+5.78,\expr+1.2) (s1_\y) {.};
       \node[block] at (9,-\y) (block\y) {$c_t$};
      \node[draw,circle, xscale=0.75, yscale=0.75] at (-\expr+3.45,-1.2) (z) {$z_{t}$};
       \node[] at (13,-\y) (input_next\y) {$\dots$};
       \draw[->, line width=0.6] (input_prev\y.east) -- (input\y);
       \draw[->, line width=0.6] (input\y) --  (s1_\y.west);
	\draw[->, line width=0.6] (f_\y.east) --(f2_\y.west);
	\draw[->, line width=0.6] (f2_\y.east) --(block\y);
	\draw[->, line width=0.6] (s1_\y.east) --(f_\y.west);
       \draw[->, line width=0.6] (block\y.east) -- (input_next\y);

    \draw[densely dashed, line width=0.3] (input\y.east) ++(0.0,0) to [out=0,in=180] ++(0.6,0.8) -- (s0_\y) -- ++(2.4,0) to [out=0,in=180] ++(0.8,-0.8);

   }
   
   \node[] at (0,-3) (h_prev) {$\dots$};
   \node[block] at (1,-3) (h) {$h_{t-1}$};
   \node[block] at (1,-4) (x) {$x_{t}$};
    \node[block] at (1,-1.5) (y) {$y_{t-1}$};
    \node[block] at (3,-1.5) (s) {$s_{t}$};
   \node[block] at (11.9,-3) (h2) {$h_{t}$};
   \node[] at (13,-3) (h_next) {$\dots$};

   \draw[->, line width=0.6] (h_prev.east) -- (h); 
   \draw[->, line width=0.6] (h2.east) -- (h_next); 
   \draw[line width=0.3] (h.east) ++(7.1,0) -- (h);  
   \draw[line width=0.3] (h.east) ++(9.45,0) -- (h2); 
   
\node[label={[label distance=0.5cm,text depth=-1ex,rotate=0]right:}] at (4.35,-0.45) {${z}_t$};
\node[label={[label distance=0.5cm,text depth=-1ex,rotate=0]right:}] at (4.95,+0.45) {$\bar{z_t}$};
\node[label={[label distance=0.5cm,text depth=-1ex,rotate=0]right:memory}] at (5.45,0.35) {};
\node[label={[label distance=0.5cm,text depth=-1ex,rotate=-90]right:control logic}] at (7.0,0.0) {};
\node[label={[label distance=0.5cm,text depth=-1ex,rotate=0]right:internal state}] at (4.2,-3.2) {};


   \draw[->-=.99, line width=0.3] (f_block0.south) ++(-0.3,-1.04) to [out=2,in=-93] (f_block0.south); 

	\draw[->, line width=0.3] (f_block0.north)  to [out=90,in=-90] (f_0);
	\draw[->, line width=0.3] (ig_block0.north)  to [out=90,in=-90] (f2_0);
   \draw[->-=.99, line width=0.3] (i_block0.south) ++(-0.3,-0.82) to [out=2,in=-93] (i_block0.south); 
   \draw[->-=.99, line width=0.3] (g_block0.south) ++(-0.4,-1.53) to [out=2,in=-93] (g_block0.south); 
	\draw[<-, line width=0.3] (o_block0.west)  to [out=180,in=0] ++(-0.72,-1.8);
   \draw[<-, line width=0.3] (oc_block0) -- ++(-0,0.49) to [out=90,in=0] ++(-0.2,0.7);
  \draw[->, line width=0.3] (i_block0.north) -- (ig_block0.south);
  \draw[->, line width=0.3] (g_block0.west) -- (ig_block0.east);
   \draw[->, line width=0.3] (o_block0.east) -- (oc_block0.west);

   \draw[->-=.94, line width=0.3] -- ([xshift=-6.066cm, yshift=-11.45em]f2_0) ++(0,0.0) -- ++(2.15,0) to [out=0,in=180] ++(0.45,1.024); 
   \draw[->-=.94, line width=0.3] -- ([xshift=-6.066cm, yshift=-11.45em]f2_0) ++(0,0.0) to [out=0,in=-90] ++(0.5,0.925)  arc(90:-90:-0.1cm) to [out=90,in=180] (s.west); 

\draw[->, line width=0.3] (h.north) -- (y.south); 
\draw[->, line width=0.3] (y.east) -- (s.west); 

   \draw[->-=.94, line width=0.3] -- ([xshift=-4.266cm, yshift=-4.25em]f2_0) ++(0,0.0) -- ++(0.3,0) to [out=0,in=180] ++(0.55,-1.51);
   \draw[->-=.999, line width=0.3] -- ([xshift=-4.266cm, yshift=-4.25em]f2_0) ++(0,0.0) -- ++(0.3,0) to [out=0,in=180] (z);
   \draw[<-=.99, line width=0.3] -- (s0_0) -- ++(0,-0.7)  arc(-90:90:-0.1cm)  to [out=270,in=0] (z);
   \draw[->-=.99, line width=0.3] -- (z.east) ++(0,0.0) -- ++(0.01,0) to [out=0,in=270] (s1_0);
   \draw[->-=.95, line width=0.3] -- (oc_block0) to [out=-0,in=180] ++(0.8,-1.8);

\end{tikzpicture}
   \caption{Sparse LSTM basic operational unit with suprisal-driven adaptive zoneout. Dashed line denotes the \emph{zoneout} memory lane.}
   \label{fig:adazo}
\end{figure}

We used the surprisal-feedback LSTM \cite{rocki2016surprisal}: 
\begin{equation}
s_{t} = {\log {p_{t-1}} \cdot x_{t}^T }
\end{equation}
Next we compute the gate activations:
\begin{equation}
f_t = \sigma({W_f \cdot x_t + U_f \cdot h_{t-1} + V_f \cdot s_{t} + b_f})
\end{equation}
\begin{equation}
i_t = \sigma({W_i \cdot x_t + U_i \cdot h_{t-1} + V_i \cdot s_{t} + b_i})
\end{equation}
\begin{equation}
o_t = \sigma({W_o \cdot x_t + U_o \cdot h_{t-1} + V_o \cdot s_{t} + b_o})
\end{equation}
\begin{equation}
{u}_{t} = tanh({W_u \cdot x_t + U_u \cdot h_{t-1} + V_u \cdot s_{t} + b_u})
\end{equation}
The zoneout rate is adaptive; it is a function of $s_t$, $\tau$ is a threshold parameter added for numerical stability, $W_y$ is a $h \rightarrow y$ connection matrix:
\begin{equation}
S_{t} = p_{t-1} - x_{t}
\end{equation}
\begin{equation}
z_{t} = min(\tau  + |S_t  \cdot W_y^T|, 1)
\end{equation}
Sample a binary mask $Z_t$ according to zoneout probability $z_t$:
\begin{equation}
Z_{t} \sim z_{t}
\end{equation}
New memory state depends on $Z_t$. Intuitively, $Z_t$ = 0 means NOP, that is dashed line in Fig. \ref{fig:adazo}.
\begin{equation}
c_{t} = (1 - f_t \odot Z_t) \odot c_{t-1} + Z_t \odot i_t \odot {u}_{t}
\end{equation}
\begin{equation}
\hat{c}_{t} = tanh(c_{t}){}
\end{equation}
\begin{equation}
h_{t} = o_t \odot \hat{c}_{t}
\end{equation}
Outputs:
\begin{equation}
y_t = W_y \cdot h_t + b_y
\end{equation}
Normalize every output unit:
\begin{equation}
p^i_t = \frac{e^{y_t^i}}{\sum_i{e^{y_t^i}}}
\end{equation}


%
%


\section{Experiments}
\subsection{Datasets}
\paragraph{Hutter Prize Wikipedia}Hutter Prize Wikipedia (also known as enwik8) dataset (Hutter, 2012). 
\paragraph{Linux}This dataset comprises approximately 603MB of raw Linux 4.7 kernel source code\footnote{ \url{http://olab.is.s.u-tokyo.ac.jp/~kamil.rocki/data/}}
\subsection{Methodology}
In both cases the first 90\% of each corpus was used for training, the next 5\% for validation and the last 5\% for reporting test accuracy. In each iteration sequences of length 10000 were randomly selected. The learning algorithm used was Adadelta with a learning rate of 0.001. Weights were initialized using the so-called Xavier initialization \cite{Glorot10understandingthe}. Sequence length for BPTT was 100 and batch size 128. In all experiments only one layer of 4000 LSTM cells was used. States were carried over for the entire sequence of 10000 emulating full BPTT. Forget bias was set initially to 1. Other parameters were set to zero. The algorithm was written in C++ and CUDA 8 and ran on GTX Titan GPU for up to 2 weeks. 

%



\begin{figure}[h]
\centering 
\includegraphics[scale=0.22]{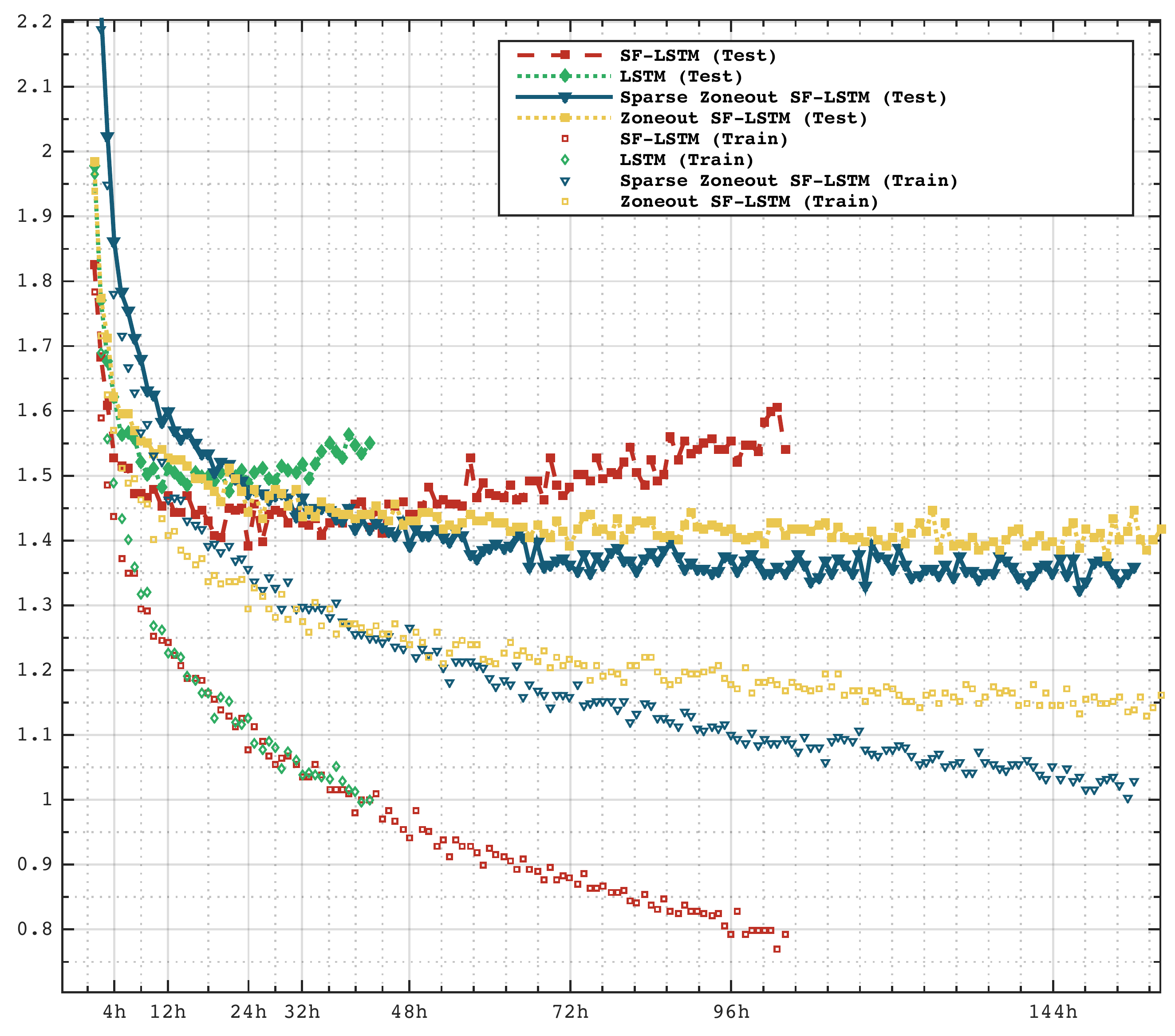}
\caption{Learning progress on enwik8 dataset}
\label{fig:learning}
\end{figure}



    \begin{figure}
        \centering
        
        \begin{subfigure}[b]{0.22\textwidth}
            \centering
            \includegraphics[width=\textwidth]{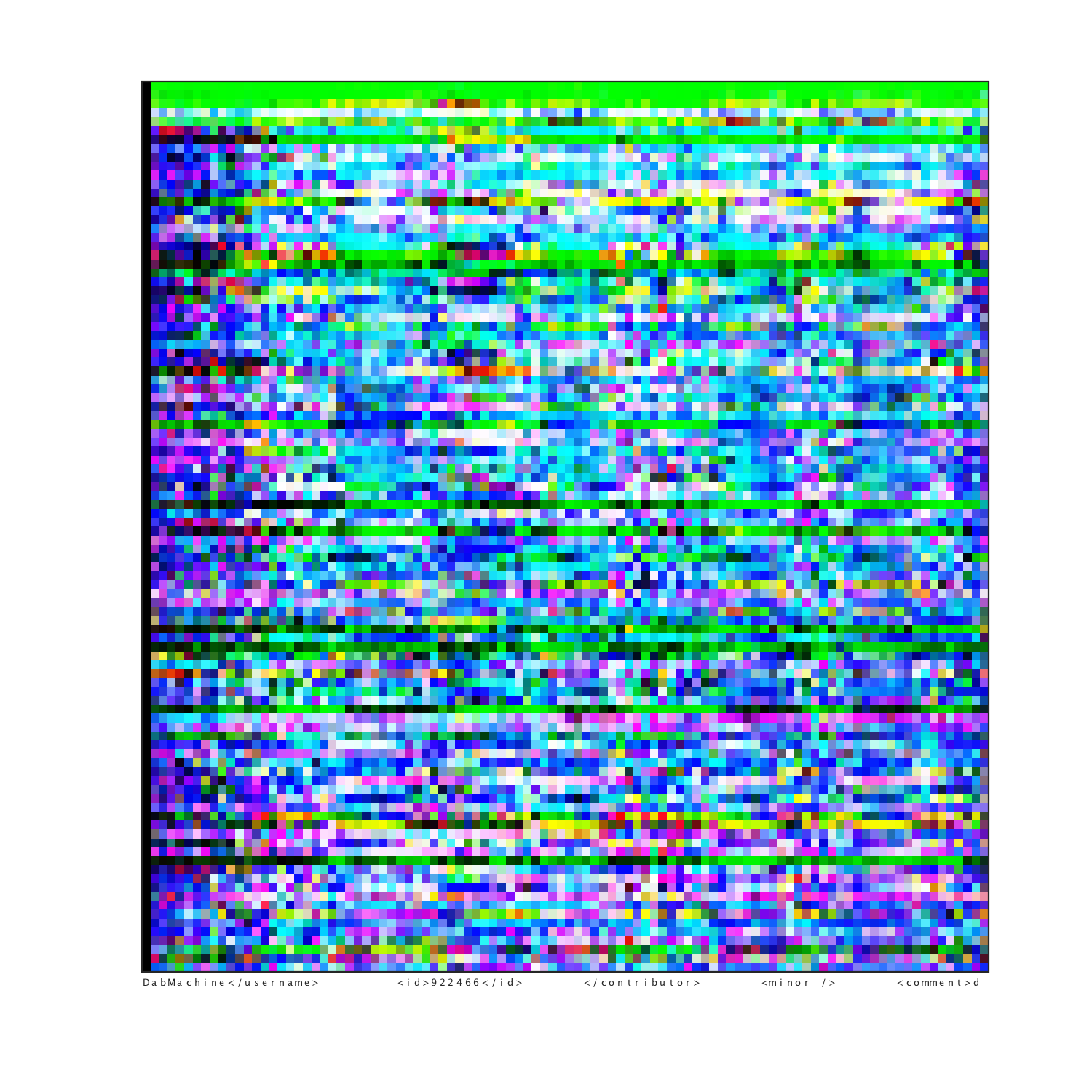}
            \caption[]%
            {{}}    
            \label{fig:nozo}
        \end{subfigure}
        \quad
        \begin{subfigure}[b]{0.22\textwidth}  
            \centering 
            \includegraphics[width=\textwidth]{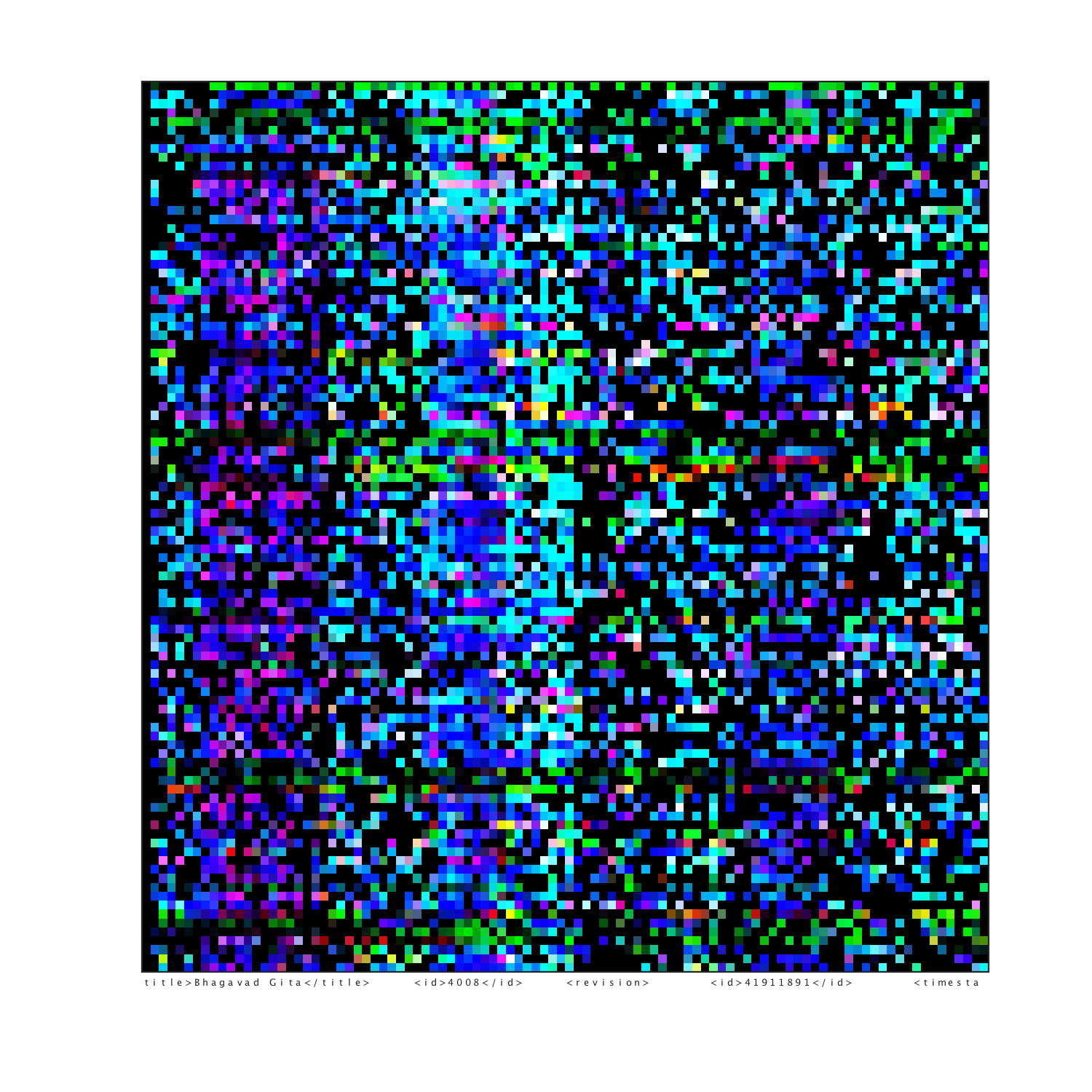}
            \caption[]%
            {{}}    
            \label{fig:zo}
        \end{subfigure}
        {{\small Colors indicate I/O gate activations: Red -- write (i), Green -- read (o), Blue - erase (f), White -- all, Black -- no operation} } 
        \vskip\baselineskip
        \begin{subfigure}[b]{0.22\textwidth}   
            \centering 
            \includegraphics[width=\textwidth]{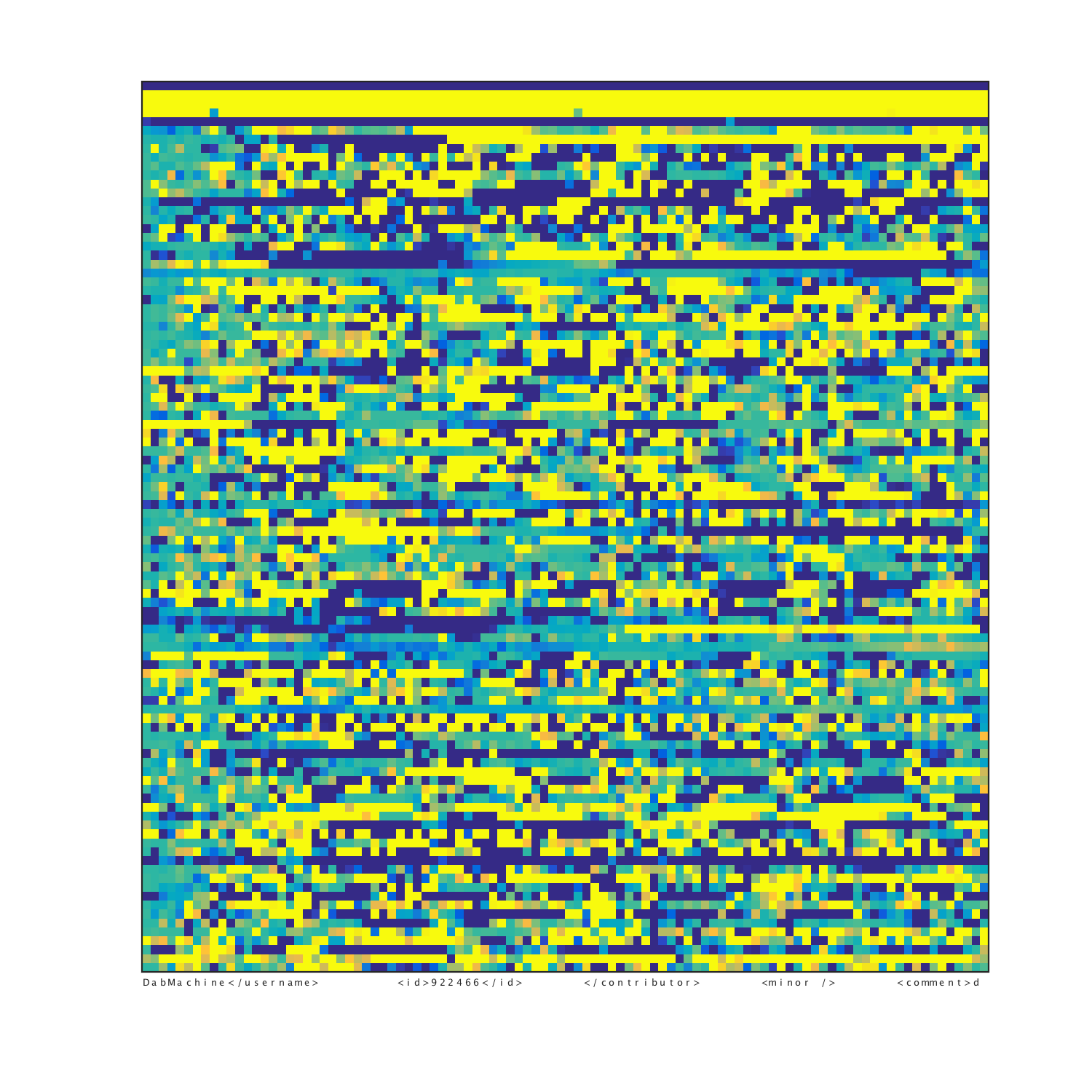}
            \caption[]%
            {{}}    
            \label{fig:mean and std of net34}
        \end{subfigure}
        \quad
        \begin{subfigure}[b]{0.22\textwidth}   
            \centering 
            \includegraphics[width=\textwidth]{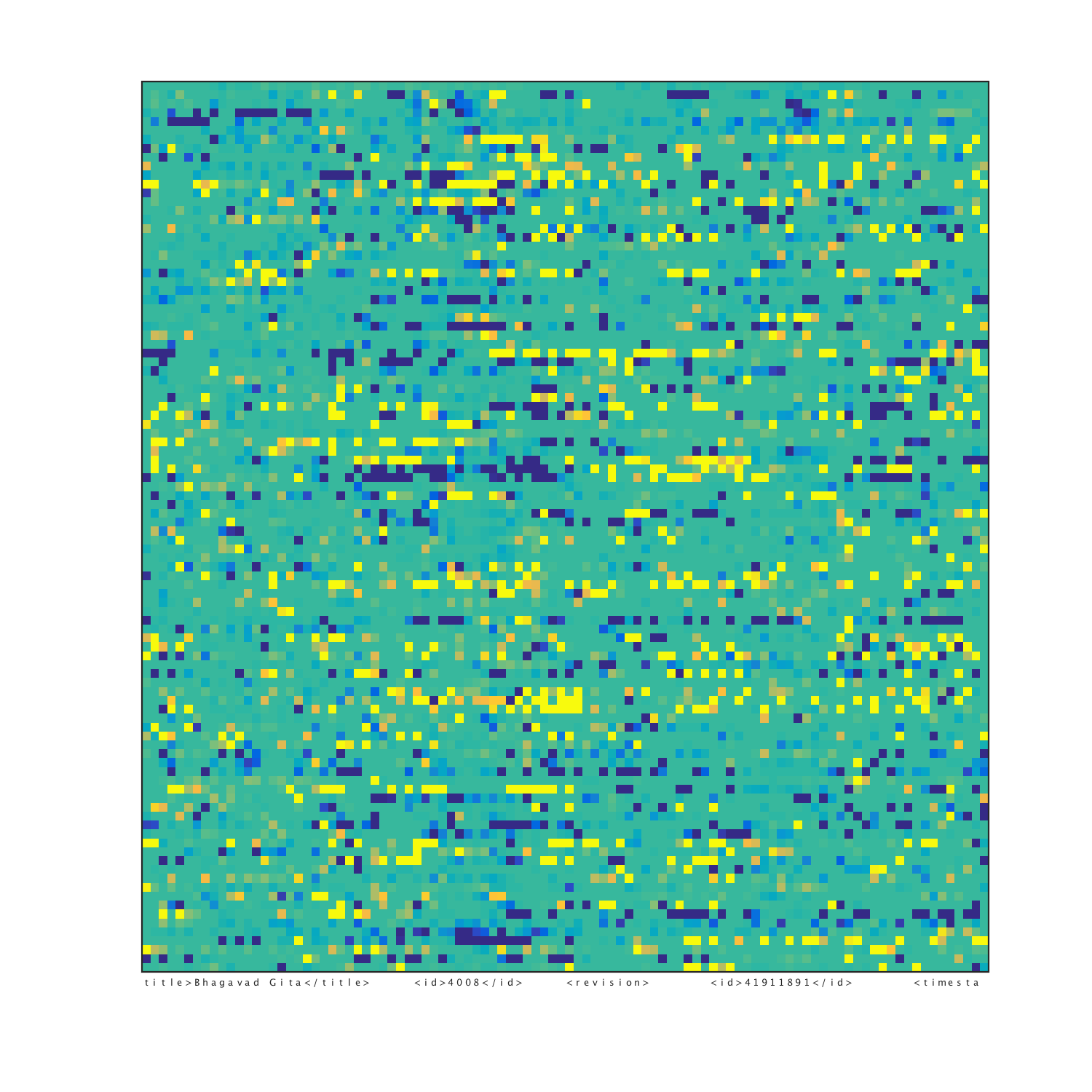}
            \caption[]%
            {{}}    
            \label{fig:mean and std of net44}
        \end{subfigure}
 	{{\small Hidden state}}  
         \vskip\baselineskip
        \begin{subfigure}[b]{0.22\textwidth}   
            \centering 
            \includegraphics[width=\textwidth]{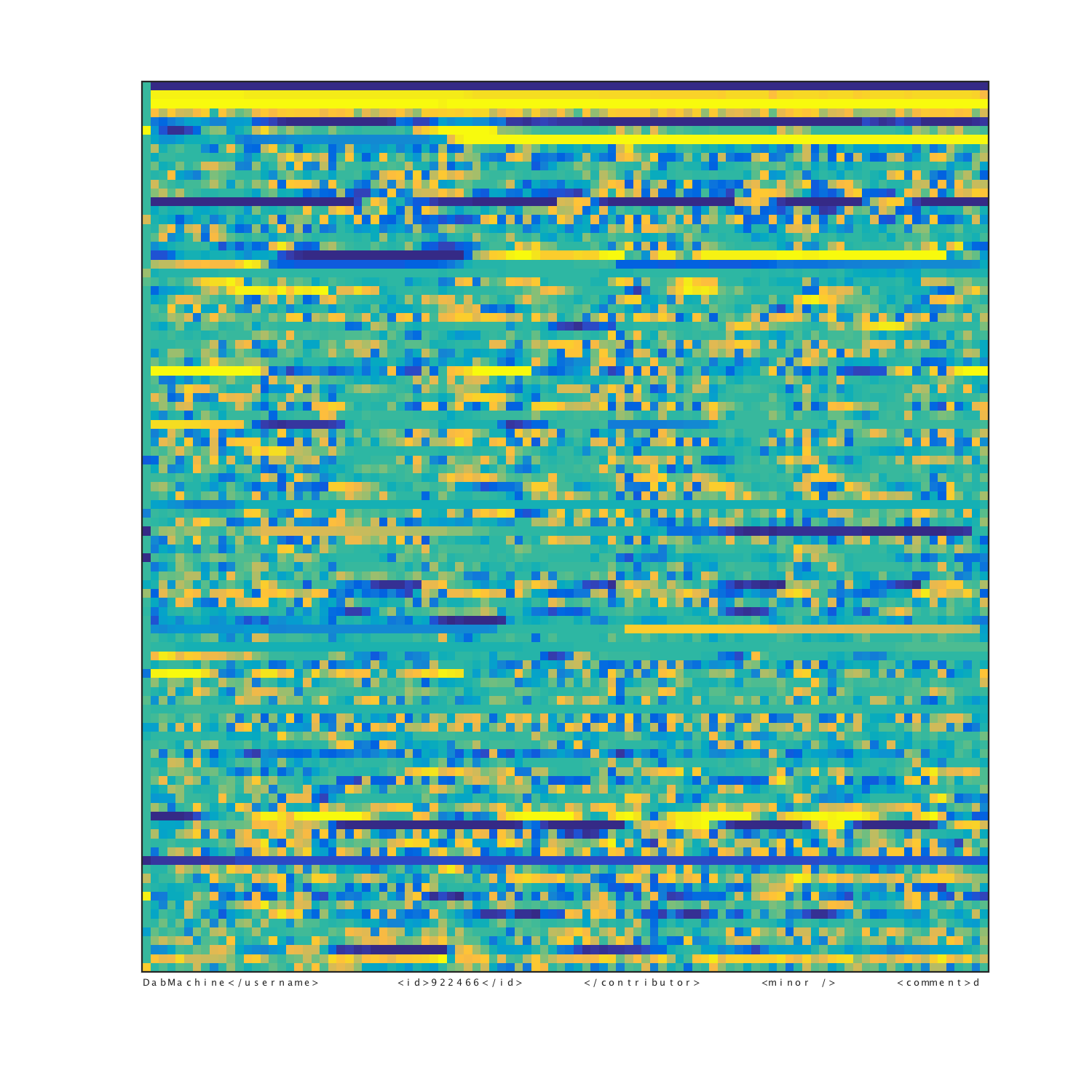}
            \caption[]%
            {{}}    
            \label{fig:mean and std of net34}
        \end{subfigure}
        \quad
        \begin{subfigure}[b]{0.22\textwidth}   
            \centering 
            \includegraphics[width=\textwidth]{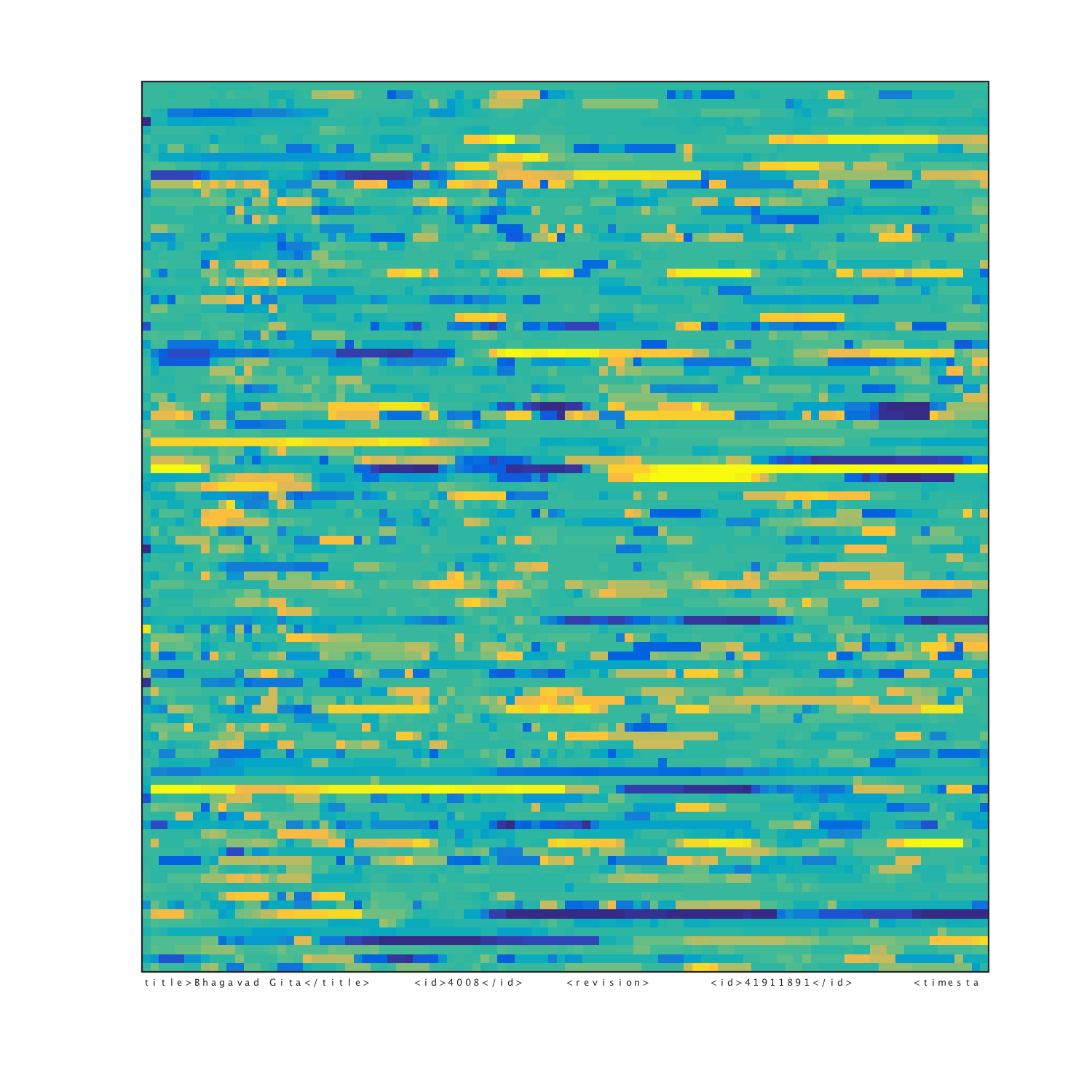}
            \caption[]%
            {{}}    
            \label{fig:mean and std of net44}
        \end{subfigure}
 	{{\small Memory cell state}}  
 \vskip\baselineskip
        \begin{subfigure}[b]{0.22\textwidth}   
            \centering 
            \includegraphics[width=\textwidth]{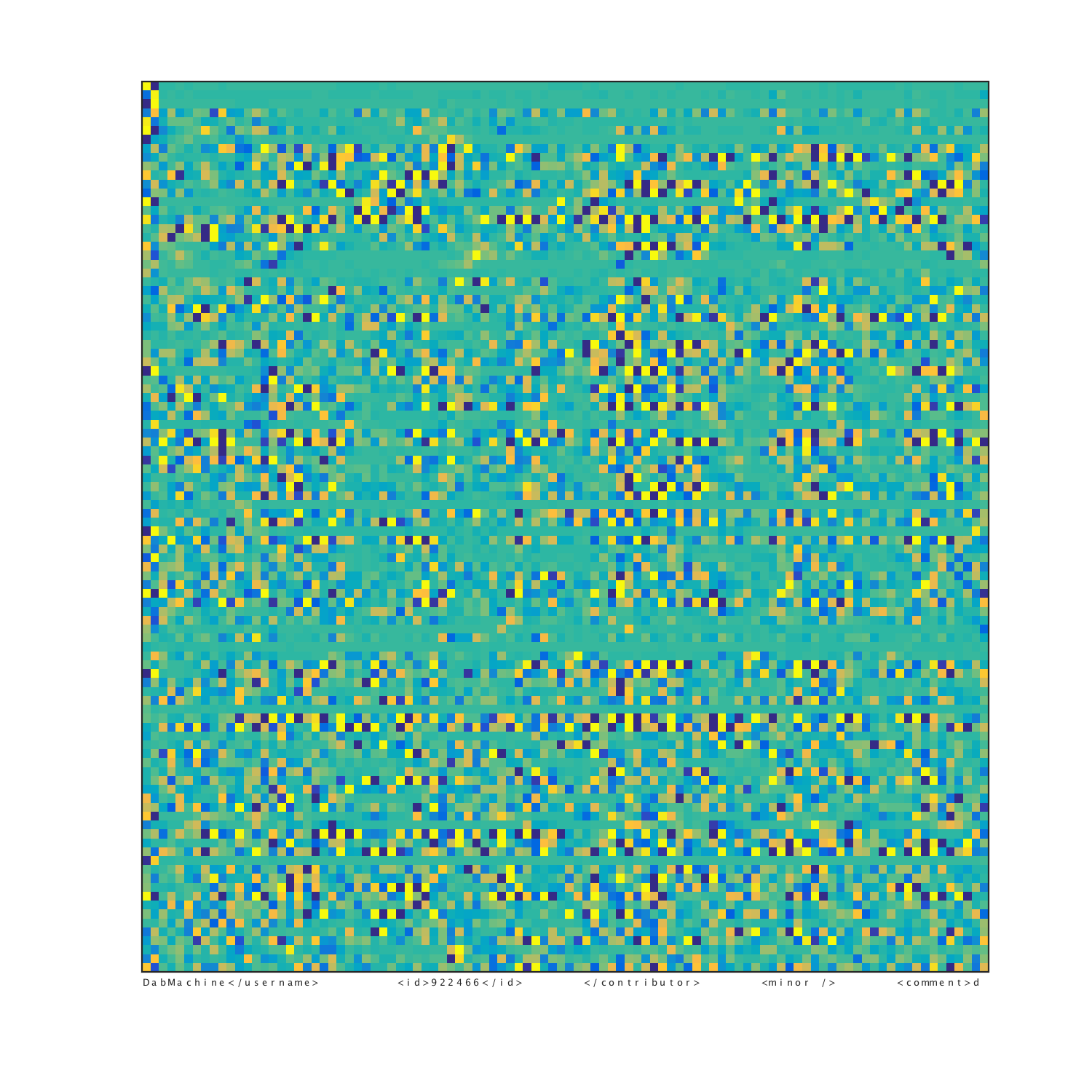}
            \caption[]%
            {{Mean: 0.27/timestep}}    
            \label{fig:mean and std of net34}
        \end{subfigure}
        \quad
        \begin{subfigure}[b]{0.22\textwidth}   
            \centering 
            \includegraphics[width=\textwidth]{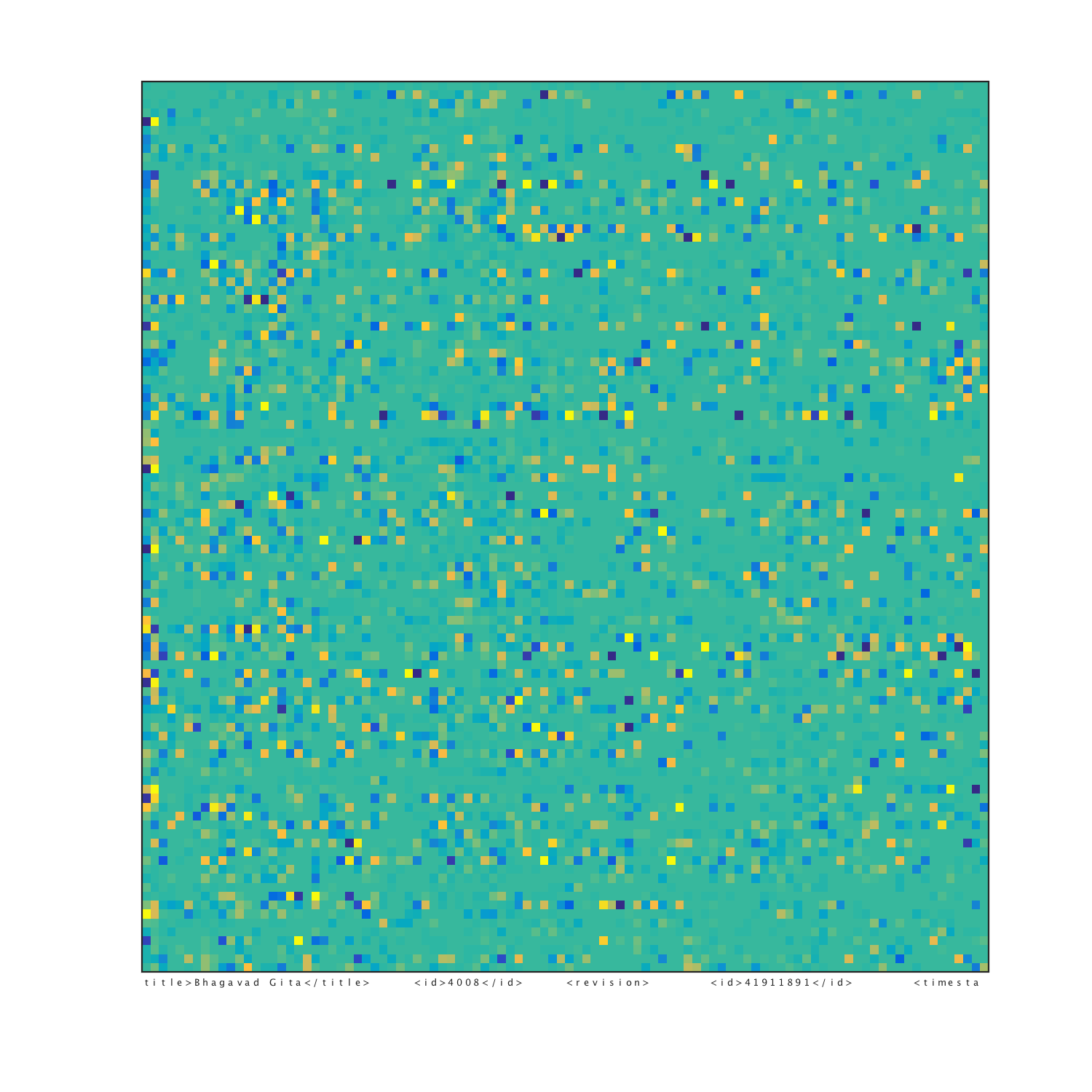}
            \caption[]%
            {{Mean: 0.092/timestep}}  
            \label{fig:mean and std of net44}
        \end{subfigure}
        {{\small Memory cell change (L1-norm)}}  
   \caption{Visualization of memory and network states in time (left to right, 100 time steps, Y-axis represents cell index); Left: without Adaptive Zoneout, Right: with Adaptive Zoneout}\label{fig:exp}
\
        \label{fig:act}
    \end{figure}


\subsection{Results and discussion}
Remark: Surprisal-Driven Feedback has sometimes been wrongly characterized as a 'dynamic evaluation' method. This is incorrect for the following reasons : 1. It never actually sees test data during training. 2. It does not adapt weights during testing. 3. The evaluation procedure is exactly the same as using standard RNN - same inputs.  Therefore it is fair to compare it to 'static' methods.
\begin{table}[h]
  \centering
  \caption{Bits per character on the enwik8 dataset (test) }
  \label{tab:bpc_enwik8}
  \begin{tabular}{rc}
    \toprule
    & BPC \\
    \midrule
    mRNN\footnotemark[1]\citep{ICML2011Sutskever_524} & 1.60  \\
    GF-RNN \citep{DBLP:journals/corr/ChungGCB15} &  1.58  \\
    Grid LSTM \citep{DBLP:journals/corr/KalchbrennerDG15} & 1.47  \\
    Layer-normalized LSTM \citep{ba2016layer} & 1.46  \\
    Standard LSTM\footnotemark[3] & 1.45 \\
    MI-LSTM \citep{DBLP:journals/corr/WuZZBS16} & 1.44 \\
     Array LSTM \citep{rocki2016recurrent} & 1.40 \\
     HM-LSTM \citep{chung2016hierarchical} & 1.40 \\
     HyperNetworks \citep{ha2016hypernetworks} & 1.38 \\
     SF-LSTM \citep{rocki2016surprisal} & {1.37} \\
     RHN \citep{DBLP:journals/corr/ZillySKS16} & 1.32  \\
    \midrule
	Surprisal-Driven Zoneout & \textbf{1.31} \\
    \midrule
    cmix v11\footnotemark[2] & {1.245} \\
    \bottomrule
  \end{tabular}
\end{table}

\begin{table}[h!]
  \centering
  \caption{Bits per character on the Linux dataset (test) }
  \label{tab:pbc_linux}
  \begin{tabular}{rc}
    \toprule
    & BPC \\
    \midrule
       SF-LSTM & {1.38} \\
       Surprisal-Driven Zoneout & \textbf{1.18} \\
    \bottomrule
  \end{tabular}
\end{table}

We observed substantial improvements on enwik8 (Table \ref{tab:bpc_enwik8}) and Linux (Table \ref{tab:pbc_linux}) datasets. Our hypothesis is that it is due to the presence of memorizable tags and nestedness in this dataset, which are ideal for learning with suprisal-driven zoneout. Patterns such as $<timestamp>$ or long periods of spaces can be represented by a single \textit{impulse} in this approach, \textit{zoning} out entirely until the end of the pattern. Without adaptive zoneout, this would have to be controlled entirely by learned gates, while suprisal allows quick adaptation to this pattern.  Fig \ref{fig:act} shows side by side comparison of a version without and with adaptive zoneout, demonstrating that in fact the dynamic span of memory cells is greater when adaptive zoneout is used. Furthermore,  we show that the activations using adaptive zoneout are in fact sparser than without it, which supports our intuition about the inner workings of the network. An especially interesting observation is the fact that adaptive zoneout seems to \textit{help} separate instructions which appear mixed  otherwise (see Fig \ref{fig:act}). A similar approach to the same problem is called Hierarchical Multiscale Recurrent Neural Networks \cite{chung2016hierarchical}. The main difference is that we do not design an explicit hierarchy of levels, instead allowing each neuron to operate on arbitrary timescale depending on its zoneout rate. Syntactic patterns in enwik8 and linux datasets are highly nested. For example ($<page>$, $<revision>$, $<comment>$,  [[:en, ..., not mentioning parallel semantic context (movie, book, history, language). We believe that in order to learn such complex structure, we need distributed representations with every neuron operating at arbitrary time scale independent of another. Hardcoded hierarchical architecture will have problems solving such a problem.

\footnotetext[2]{Best known compressor: http://mattmahoney.net/dc/text.html}
\footnotetext[3]{our implementation}
\section{Summary}


The proposed surprisal-driven zoneout appeared to be a flexible mechanism to control the activation of a given cell. Empirically, this method performs extremely well on the enwik8 and linux datasets. 

\section{Further work}

We would like to explore variations of suprisal-driven zoneout on both state and cell. Another interesting direction to pursue is the connection with sparse coding - using suprisal-driven zoneout, the LSTM's cell contents are more sparsely revealed through time, potentially resulting in information being used more effectively.


\section*{Acknowledgements}
This work has been supported in part by the Defense Advanced Research Projects Agency (DARPA).

\bibliography{slstm}

\end{document}